\documentclass[conferemce]{IEEEtran}
\usepackage{graphicx}
\usepackage{epstopdf}
\usepackage{amsthm}
\usepackage{epsfig}
\usepackage{latexsym}
\usepackage{amsfonts}
\usepackage{here}
\usepackage{rawfonts}
\usepackage[latin1]{inputenc}
\usepackage[T1]{fontenc}
\usepackage{calc}
\usepackage{capitalgreekitalic}
\usepackage{url}
\usepackage{enumerate}
\usepackage{color}
\usepackage[tbtags]{amsmath}
\usepackage{amssymb}
\usepackage{upref}
\usepackage{epic,eepic}
\usepackage{times}
\usepackage{dsfont}
\usepackage{comment}
\usepackage{cite}
\usepackage{bbm}
\usepackage{bm}
\usepackage{amsmath}
\usepackage{booktabs}
\usepackage{makecell}
\usepackage{csquotes}
\usepackage{multirow}
\usepackage{subfigure}

\usepackage{multirow}
\usepackage{array}

\allowdisplaybreaks[4]

\usepackage{dsfont}

\newlength{\aligntop}
\newlength{\alignbot}
 \makeatletter
\renewenvironment{align}{%
  \vspace{\aligntop}
  \start@align\@ne\st@rredfalse\m@ne
}{%
  \math@cr \black@\totwidth@
  \egroup
  \ifingather@
    \restorealignstate@
    \egroup
    \nonumber
    \ifnum0=`{\fi\iffalse}\fi
  \else
    $$%
  \fi
  \ignorespacesafterend%
  \vspace{\alignbot}\par\noindent
} \makeatother

\IEEEoverridecommandlockouts

\usepackage{algorithm}
\usepackage{algorithmic}

\makeatletter
\newcommand\semihuge{\@setfontsize\semihuge{19.3}{25}}
\makeatother

\makeatletter
\newcommand\semismall{\@setfontsize\semihuge{12.4}{15}}
\makeatother

\usepackage{subfigure}

\begin{document}
\title{\huge 
Attention-based UNet enabled Lightweight Image Semantic Communication System  over Internet of Things
}
\author{\IEEEauthorblockN{
Guoxin Ma,
Haonan Tong, 
Nuocheng Yang, 
and Changchuan Yin}\\
\vspace{0.3cm}
\IEEEauthorblockA{\small Beijing Laboratory of Advanced Information Network, Beijing University of Posts and Telecommunications, Beijing, China\\
\thanks{This work was supported in part by Beijing Natural Science Foundation under Grant L223027, the National Natural Science Foundation of China under Grant 61671086, China 973 Program under Grant 2009CB320407, in part by BUPT Excellent Ph.D. Students Foundation under Grant CX2021114, and China Scholarship Council.}
}
Emails: 
\{guoxinma, hntong, yangnuocheng, and ccyin\}@bupt.edu.cn

}
\maketitle

\pagestyle{empty}  
\thispagestyle{empty} 
\begin{abstract}

This paper studies the problem of the lightweight image semantic
communication system that is deployed on Internet of Things (IoT) devices. In the considered system model, devices must use semantic communication techniques to support user behavior recognition in ultimate video service with high data transmission efficiency. However, it is computationally expensive for IoT devices to deploy semantic codecs due to the complex calculation processes of deep learning (DL) based codec training and inference. To make it affordable for IoT devices to deploy semantic communication systems, we propose an attention-based UNet enabled lightweight image semantic communication (LSSC) system, which achieves low computational complexity and small model size. In particular, we first let the LSSC system train the codec at the edge server to reduce the training computation load on IoT devices. Then, we introduce the convolutional block attention module (CBAM) to extract the image semantic features and decrease the number of downsampling layers thus reducing the floating-point operations (FLOPs). Finally, we experimentally adjust the structure of the codec and find out the optimal number of downsampling layers. Simulation results show that the proposed LSSC system can reduce the semantic codec FLOPs by 14$\%$, and reduce the model size by 55$\%$, with a sacrifice of 3$\%$ accuracy, compared to the baseline. Moreover, the proposed scheme can achieve a higher transmission accuracy than the traditional communication scheme in the low channel signal-to-noise (SNR) region. 

\end{abstract}

\begin{IEEEkeywords}
Semantic communication, Image segmentation, Attention Mechanism, Internet of Things
\end{IEEEkeywords}

\IEEEpeerreviewmaketitle
\vspace{-1.5em}
\section{Introduction}

Semantic communication is a potentially vital technology to enable efficient data transmission for 6G intelligent connection, with the reduced amount of transmission data over ubiquitous intelligence applications \cite{strinati20216g,veedu2022toward,gong2023adaptive}. To deploy a semantic communication system, Internet of Things (IoT) devices need Deep Learning (DL) based codecs for real-time semantic extraction and signal processing. However, training the codec on IoT devices and deploying the DL-based codec is computationally expensive, which can bring huge overhead for IoT devices with limited energy. 
 Therefore, designing a lightweight semantic codec to reduce the costs of DL models on IoT devices is significant for deployment on IoT devices that support ubiquitous intelligence.

Existing works such as \cite{xie2021deep,weng2021semantic,tong2021federated} have studied the efficient transmission of semantic communication for various data types over the Internet of Things. The authors in \cite{xie2021deep} proposed a DL-based semantic communication system, named DeepSC, for text transmission, which is more robust to channel variation compared with the traditional communication system. The work in \cite{weng2021semantic} proposed a DL-enabled semantic communication system for speech signals (DeepSC-S) to improve the transmission efficiency significantly. Moreover, in \cite{tong2021federated}, the authors proposed a federated learning-based audio semantic communication system over wireless networks, which can converge effectively compared to a traditional coding scheme. However, the works in \cite{xie2021deep,weng2021semantic,tong2021federated} have left the field of image semantic communication, which is challenging due to the complexity of image data structure and various downstream tasks.

The prior arts \cite{xie2020lite,du2023yolo} have studied lightweight image semantic communication along with the development of image segmentation techniques. The work in \cite{xie2020lite} proposed a lightweight distributed semantic communication system by pruning the model redundancy and decreasing the weight resolution. The authors in \cite{du2023yolo} developed a semantic communication framework that reduces the computational complexity to run on edge devices with limited power. 
However, now that the works in \cite{xie2020lite,du2023yolo} introduce lightweight semantic codecs. When using a lightweight semantic codec, the model accuracy inescapably gets worse. To solve this, in \cite{li2022attention}, the authors designed a feature compression module based on the channel attention module that selects the most important features. Furthermore, the work in \cite{li2022segmentation} proposed a combination of attention mechanism and dilated convolutions, to improve the segmentation performance of the image encoder. The authors in \cite{huang2023channel} proposed an efficient attention mechanism to learn the dynamic distribution of attention weights in both channel and spatial dimensions. However, most of these works \cite{li2022attention,li2022segmentation,huang2023channel} utilized the attention mechanism to improve the accuracy without considering the model size. Therefore, it is necessary to develop a lightweight semantic communication system considering both high accuracy and small model size.

\begin{figure*}[htbp]
 	\includegraphics[width=16cm,height=5cm]{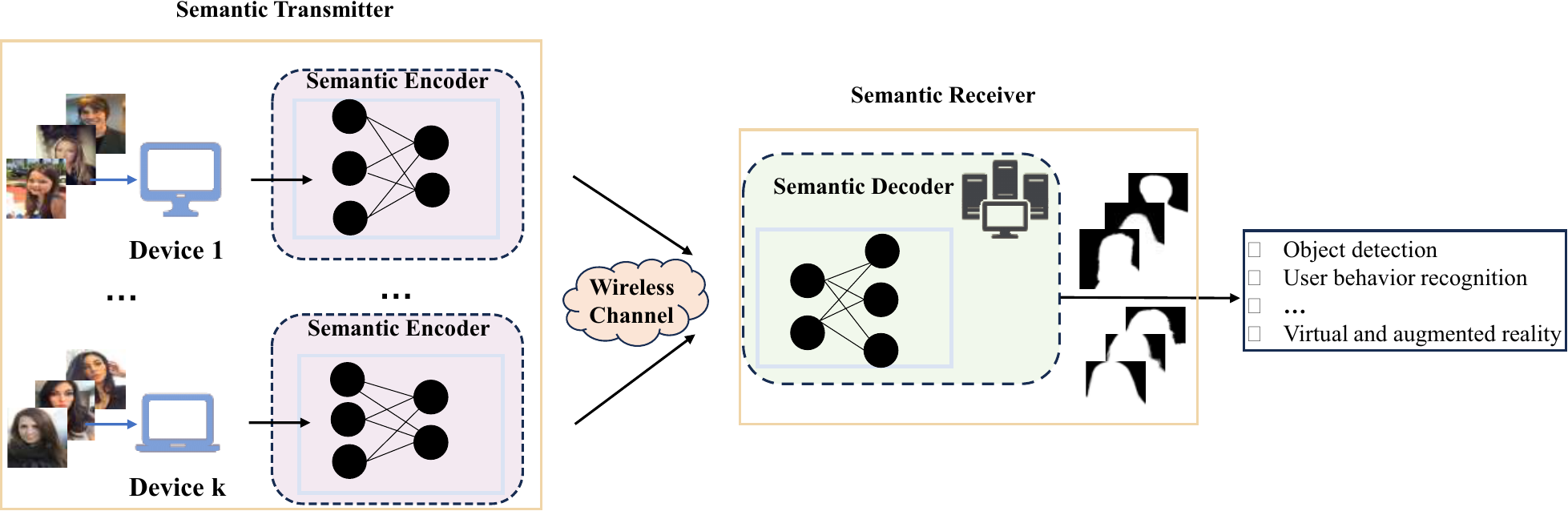}
	\caption{\centering{Architecture of the LSSC system.}}
	\label{fig2}
\end{figure*}

The main contribution of this paper is a novel image semantic segmentation communication system that can be deployed on IoT devices with low floating-point operations (FLOPs) and small model size while ensuring high segmentation performance. Our key contributions include the following.
\begin{itemize}
    \item We develop a lightweight image semantic communication (LSSC) system that transmits the trained semantic codec and semantic features between the edge server and IoT devices. The semantic codec consists of a semantic encoder that extracts the image data at the transmitter and a decoder that at the receiver decodes the image information, respectively.
    \item We design a lightweight attention-based UNet enabled network architecture with low computational complexity and small model size by decreasing the number of downsampling layers. Meanwhile, we introduce the convolutional block attention module (CBAM), which utilizes the attention module consisting of spatial and channel attention modules to evaluate the importance of individual pixels and channels for image segmentation, respectively. 
    \item Simulation results show that the proposed LSSC system can reduce the FLOPs by 14$\%$ and model size by 55$\%$ with only a 3$\%$ mean Intersection over Union (mIoU) loss compared to traditional UNet. Moreover, our proposed model achieves higher mIoU than the baseline using the traditional method under different wireless channel conditions, especially in the low channel signal-to-noise (SNR) region. 
\end{itemize}

The remainder of this paper is organized as follows. The system model and problem formulation are presented in Section \uppercase\expandafter{\romannumeral2}. In section \uppercase\expandafter{\romannumeral3}, we give a detailed description of the proposed lightweight semantic encoder and decoder. The simulation results are presented and analyzed in Section \uppercase\expandafter{\romannumeral4}. The conclusion is drawn in Section \uppercase\expandafter{\romannumeral5}.

\vspace{-1.1em}
\section{System Model And Problem Formulation}
 We consider an IoT network to deploy the LSSC system as shown in Fig. 1. The LSSC system consists of multiple devices and an edge server. To support wireless data transmission, IoT devices extract the image semantic information and transmit it. Then, the image semantic segmentation information obtained at the receiver will be used to support the downstream tasks, such as object detection, user behavior recognition, and virtual reality. Since the trained codec is deployed on IoT devices with finite memory, we must design a semantic segmentation codec that is lightweight enough to be deployed on devices. Moreover, since the computational capability of the device is limited, the semantic codec must be trained on the server with huge enough computational capability. As shown in Fig. 2, the edge server trains the codec and multiple devices transmit the image semantic features to the edge server. In this scenario, the overall training and inference procedure of the LSSC system is given as follows:
 \begin{enumerate}
     \item The edge server trains the semantic codec, which includes an encoder and a decoder.
     \item The Edge server transmits the trained semantic encoder to each IoT device.
     \item Each device collects the images and extracts image semantic features which are transmitted to the edge server.
     \item The decoder at the edge server obtains the image segmentation which can be utilized for the downstream tasks. 
 \end{enumerate}

\begin{figure}[t]
\centering
\setlength{\belowcaptionskip}{-0.45cm}
\includegraphics[width=7.5cm]{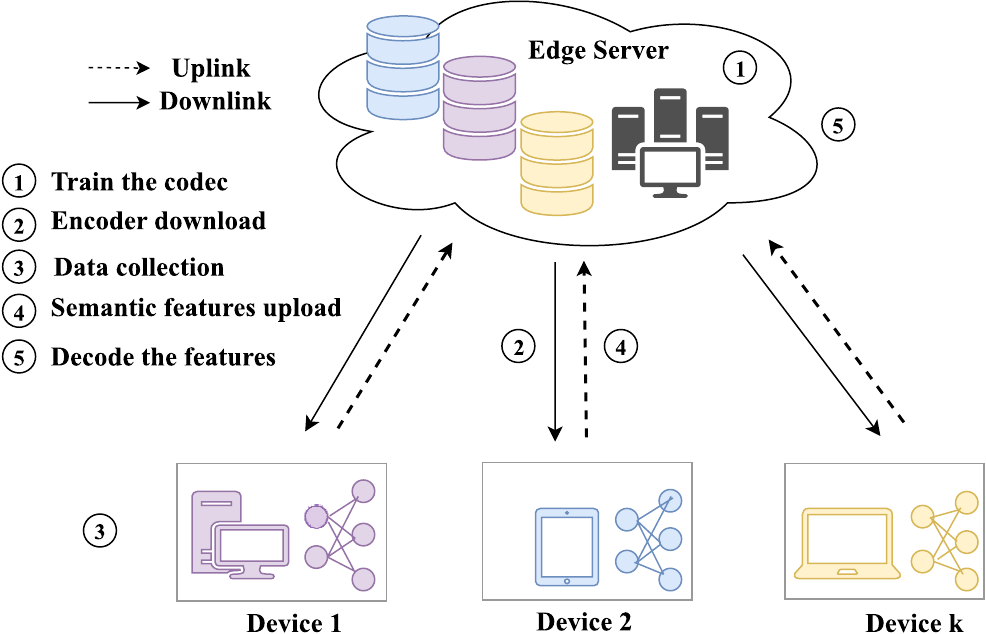}
\centering
\caption{The framework of semantic communication for networks.}
\label{fig2}
\vspace{-0.4cm}
\end{figure}

\vspace{-1.5em}
\subsection{LSSC Encoder}
The LSSC encoder extracts the different levels of semantic features from the input image data. The processed image is put into the proposed encoder to extract the semantic features. Then, the semantic features are sent through the wireless channel. 

We assume that the input image of the LSSC system is $ \boldsymbol{S} \in \mathbb{R}^{H \times W \times C}$, with  $H, W$, and $C$ being the image width, height, and the number of channels, respectively. We simplify the LSSC encoder parameter as $\alpha$, thus, the relationship between the transmitted semantic features $\boldsymbol{x}$ and the input image $\boldsymbol{S}$ can be given by
\begin{equation}
    \begin{aligned}
        \boldsymbol{x}=R_\alpha(\boldsymbol{S}),
    \end{aligned}
\end{equation}
where $\boldsymbol{x}=R_\alpha(\cdot)$ indicates the function of the semantic encoder.

\subsection{Wireless Channel}
When transmitted over a wireless channel, the encoded semantic features will suffer channel fading and noise. We consider the condition of the wireless channel, where the received semantic features $\boldsymbol{y}$ at the LSSC decoder can be given by
\begin{equation}
\begin{split}
    \boldsymbol{y} =h \cdot\boldsymbol{x}+\boldsymbol{n},
\end{split}
\end{equation}
where $h$ represents the channel gain between the transmitter and the receiver, and $\boldsymbol{n} \sim \mathcal{N} \lvert(0, \sigma_n^2 \lvert \boldsymbol{I})$ is additive white Gaussian noise (AWGN) with variance $\sigma_n^2$ being noise variance and $\boldsymbol{I}$ being identity matrix.
\vspace{-1.1em}
\subsection{LSSC Decoder}
The LSSC decoder is used to recover the received semantic feature $\boldsymbol{y}$. The decoded image segmentation is $\boldsymbol{\widehat{S}}$. The decoder needs to get the image segmentation $\boldsymbol{\widehat{S}}$ with the same height and width of the input image.  $\boldsymbol{\widehat{S}}$ classifies the pixels in $\boldsymbol{S}$ into $N$ categories with $N$ being the number of object categories concerned in $\boldsymbol{S}$ (portraits and background, etc.). To this end, the association between the output image segmentation  $\boldsymbol{\widehat{S}}$ and the received semantic features $\boldsymbol{y}$ can be given by 
\begin{equation}
 \boldsymbol{\widehat{S}}=R_\beta^{-1}(\boldsymbol{y}),
\end{equation}
where $R_\beta^{-1}(\boldsymbol{\cdot})$ is the semantic decoder network with parameter $\beta$ and the superscript represents the inverse function operation.
\vspace{-2.3em}
\subsection{LSSC Objective}
The objective of the LSSC system is to decode the image segmentation of the image containing portraits at the receiver. Since the LSSC system focuses on semantic-level image recovery, we use the mIoU metric rather than the bit error rate in traditional communications to evaluate the system performance, which is given by
\begin{equation}
\begin{aligned}
  mIoU=\frac{1}{N}\sum\limits_{i = 1}^N \frac{P\cap G}{P \cup G} ,
\end{aligned}
\end{equation}
where $P$ is the set of pixel regions predicted by the LSSC decoder for a specific category object, and $G$ is the actual set of pixel regions for this category object.

To compress the image semantic segmentation models for transmission and implementation on devices, we propose a lightweight image semantic segmentation model. Generally, the DL model training and inference run upon FLOPs, which means that the input and output require a large number of computation cycles. The number of FLOPs required by model feedforward measures the computation complexity of the model. During forward propagation in a model, operations such as convolution, pooling, BatchNorm, ReLU, and Upsample, etc., are performed, with convolution being the most computationally intensive. Therefore, we mainly consider the decrease in convolution operation for low FLOPs. In order to deploy the efficient semantic codec on IoT devices, we need to reduce the FLOPs. The number of FLOPs in LSSC codec is defined as $G$, which can be given by
\begin{equation}
    G =\sum_{l=1}^{L} l \times W_{l} \times H_{l} \times C_{l}^{\mathbf{in}} \times C_{l}^{\mathbf{out}} \times (K_{l}^2+1),
\end{equation}
where $W_l$ and $H_l$ are the width and height of the input feature map of one downsampling or upsampling layer $l$, respectively. $K_{l} $ is the kernel size. $C_{l}^{\mathbf{in}}$ and $C_{l}^{\mathbf{out}}$ are the number of channels of the input and output feature maps of one layer, respectively. $L=L_{en}+L_{de}$ is the total number of convolutional and deconvolutional layers, with $L_{en}$ being the number of convolutional layers of LSSC encoder and $L_{de}$ being the number of deconvolutional layers of LSSC decoder.
Meanwhile, the performance of a codec improves as the number of parameters increases. However, high computational complexity and model size results in high energy consumption. Given the finite memory of devices, a model with a smaller size is favored for the advantages, including improved memory conservation, enhanced operational efficiency, and faster calculations. In the LSSC system, the model size of codec $M$ is given by
\begin{equation}
     M = \sum_{l=1}^{L} l \times C_{l}^{\mathbf{in}} \times C_{l}^{\mathbf{out}} \times K_{l}^2.
\end{equation}
Moreover, in order to exploit the trade-off between the model accuracy and model complexity, we design a lightweight model for the semantic communication system. Assuming the LSSC system, we denote the segmentation metric achieved by the LSSC codec as $Q$, which is given by
\begin{equation}
\begin{split}
    Q &=\lambda \cdot mIoU-\mu \cdot G- \nu \cdot M,
\end{split}
\end{equation}
where $\lambda$, $\mu$, and $\nu$ are coefficients that characterize the importance of accuracy, model computational complexity, and model size. 

The objective of the LSSC system is to maximize the mIoU, while with lower FLOPs and smaller model size under the proposed scenario. Therefore, we formulate the following optimization problem. The expression can be given by
\begin{equation}
   \max_{G,M} Q,
\end{equation}
\begin{align}
{\rm{s.t.}} 
    \quad & G \leq g, \tag{\theequation a}\\
    & M \leq m, \tag{\theequation b}
\end{align}
where the $g$ and $m$ are the FLOPs and model size of the baseline of the existing semantic codec. From (5), (6), and (7), we can see that the $L$, $C_{l}^{\mathbf{in}}$, $C_{l}^{\mathbf{out}}$ and $K_l$ determine the $G$ and $M$, thus determining the optimization objective $Q$. In order to reduce the FLOPs and model size, we decrease the number of $L$ to make the $G$ and $M$ smaller. Since compressing the semantic segmentation model makes it possible to run the DL model on IoT devices, it is necessary to design the architecture of the LSSC encoder and the LSSC decoder. To this end, we introduce an attention-based model to extract the image semantic information.


\vspace{-1.2em}
\section{LSSC Encoder And Decoder Design}

To solve the problem (8), we design the architecture of the neural network to optimize the stated objective. We first propose an improved network architecture to extract image information with high efficiency. Then, we provide a detailed exposition of the LSSC encoder and decoder based on UNet, along with the improvements made to it and we utilize the CBAM module to improve the extraction performance. Subsequently, we introduce the training strategy of our proposed LSSC system.  
\vspace{-1.2em}
\subsection{UNet based Semantic Codec Architecture}
For semantic segmentation tasks, the detailed high-level segmentation feature information is valuable. Based on the symmetric encoder-decoder structure of the UNet, with a cross-layer feature reconstruction module, the important high-level feature information can be preserved. Meanwhile, skip-connection can provide multi-scale and multi-level information for image segmentation, thus obtaining a more precise segmentation performance. Most importantly, UNet is better than simply using the codec framework, where much detailed information that is helpful is compressed and lost during the coding process \cite{ronneberger2015u}. Therefore, we determined to develop a lightweight model based on UNet.

Based on the UNet symmetric encoder-decoder structure, from the expressions (5) and (6), we can see that when the number of downsampling and upsampling operations decreases, it consequently influences the number of convolutional layers, resulting in changes to both FLOPs and model size. Since the computational capability of the IoT device is limited, UNet imposes a significant computational burden on IoT devices. Therefore, we need to reduce the computational complexity by ignoring some semantic features in order to lessen the burden on IoT devices. 
Furthermore, in order to address the issue of class ambiguity resulting from the reduction of high-level features, the CBAM module is employed in the semantic encoder, enabling adaptive recalibration of feature mappings.

\begin{figure}[t]
\centering
\setlength{\belowcaptionskip}{-0.45cm}
\includegraphics[width=8.5cm]{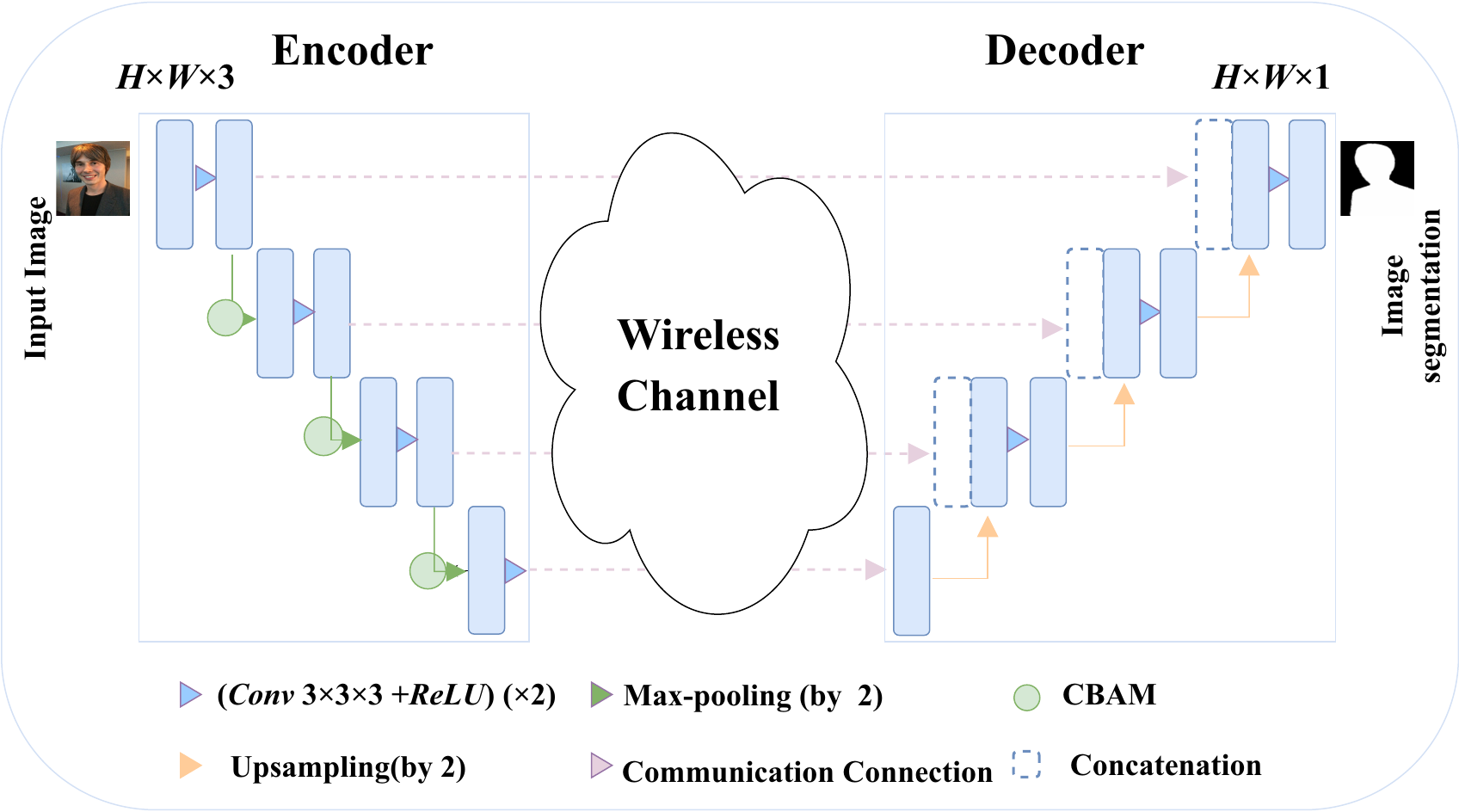}
\centering
\caption{UNet based Semantic Codec Architecture in LSSC system.}
\label{fig2}
\end{figure}
The proposed CBAM-based network architecture is shown in Fig. 3. From Fig. 3, we can see that the proposed network architecture is composed of a semantic encoder, wireless channel, and semantic decoder, which extract semantic information and recover semantic features, respectively. To extract the important image data information in the downsampling process, the CBAM module is employed following each downsampling layer while ignoring useless features. 
\vspace{-1.2em}
\subsection{Module Enhancement Methods}
Fig. 4(a) shows the CBAM module in the semantic encoder of the LSSC system. The CBAM module can assign different weights to different channels or regions in space, thereby helping the codec to focus on extracting more important information.
CBAM module is proposed as a lightweight attention module, which can infer attention maps along two separate dimensions, channel and space. Subsequently, we describe two dimensions of the CBAM module, the channel attention module, and the spatial attention module, respectively. Since CBAM is a lightweight general-purpose module, it can be seamlessly integrated into any network architecture with negligible overhead, and it can be end-to-end trainable\cite{woo2018cbam}.


\begin{figure}[t]  
	\centering
	\subfigure[CBAM Module]{
		\includegraphics[width=7.5cm]{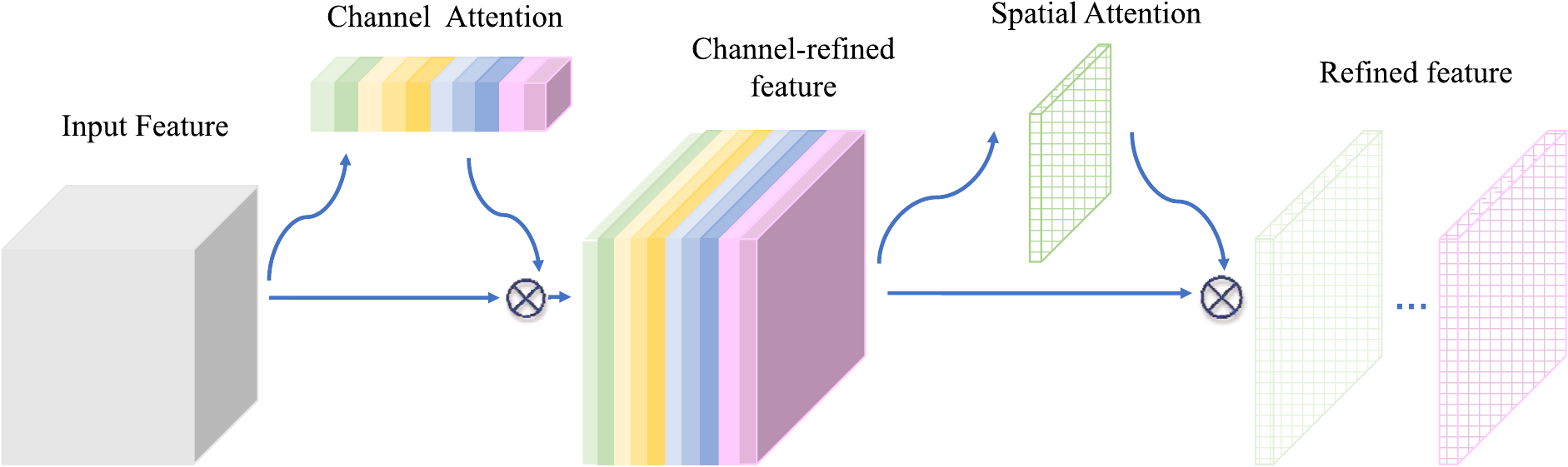}
	}
	
	\subfigure[Channel Attention Module]{
		\includegraphics[width=7.5cm]{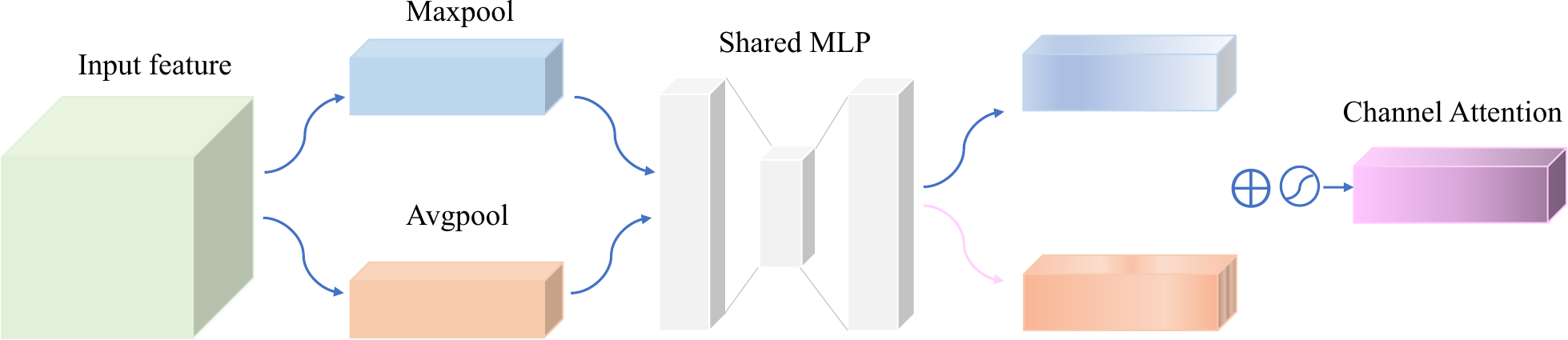}
	}
 
	\subfigure[Spatial Attention Module]{
		\includegraphics[width=7.5cm]{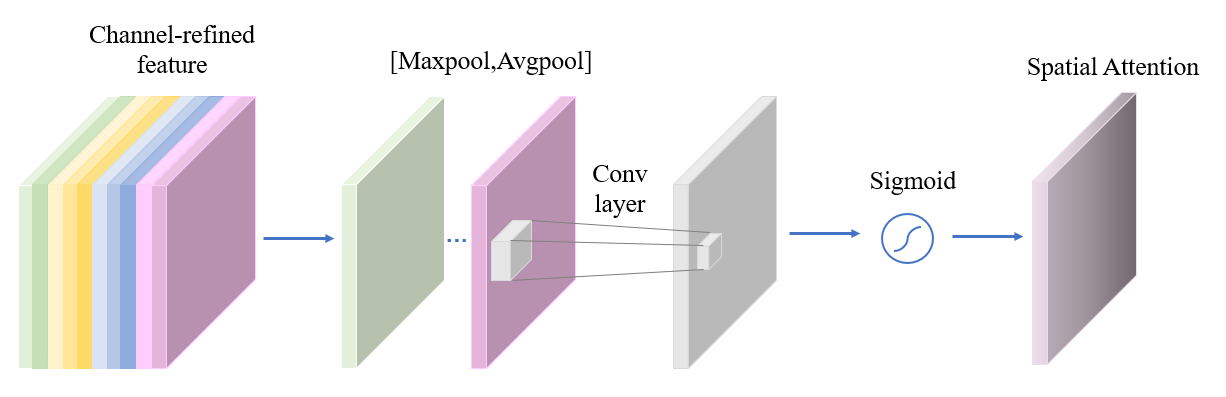}
	}

	\caption{Comparisons of different attention modules}
	\label{Fig: O2O_RF_MCC3}	
\end{figure}

The channel attention module focuses on which channels play an important role in determining the final inference results of the LSSC codec, i.e., it selects the channels that are decisive for prediction. As shown in Fig. 4(b), the input feature map $\boldsymbol{F} \in \mathbb{R}^{H \times W \times C}$ is initially fed into both maximum pooling and average pooling operations along the channel dimension. The outputs of the pooling operations are then fed into a multilayer perceptron (MLP) layer, followed by element-wise summation and activation using a sigmoid function to generate the channel attention weight map $M_{c} \in \mathbb{R}^{H \times W \times C}$. The detailed process is described as
\begin{equation}
\begin{split}
     M_{c}(F) &=\sigma(MLP(Avgpool(F))+MLP(Maxpool(F))) \\
             &=\sigma(W_1(W_0(F_{avg}^c))+W_0(F_{max}^c))),
\end{split}
\end{equation}
where $\sigma$ indicates the sigmoid function, MLP represents the multilayer perceptron, and Avgpool and Maxpool denote the pooling operations. The channel attention weight map $M_{c}$ will be used as the input of the spatial attention module. $F_{avg}^c$ and $F_{max}^c$ represent the feature after global avgpooling and maxpooling layer, $W_0$ and $W_1$ denote the two-layer weights of the MLP. The channel attention weight map $M_{c}$ will be used as the input of the spatial attention module.

The spatial attention module is designed to identify which pixels in an RGB image are most informative for making classification\cite{guo2022segnext}. This is achieved by using a similar mechanism as the channel attention module but applied at the pixel level, yielding a pixel-level attention weight map to represent the importance of each pixel. The implementation is illustrated in Fig. 4(c), where the selected informative channels tensor $F'$ is first subjected to a channel-based maximum and mean pooling operation, and then two outputs are concatenated and passed through a 7x7 convolutional layer. Finally, the resulting feature map is activated using the sigmoid function to generate the spatial attention weight map $M_{s}(F') \in \mathbb{R}^{H \times W \times C}$, which is given by 
\begin{equation}
\begin{split}
    M_{s}(F')&=\sigma(f^{7\times7}([AvgPool(F');MaxPool(F')]))\\
        &= \sigma(f^{7\times7}([F_{avg}^s;F_{max}^s])),
\end{split}
\end{equation}
where $f^{7\times7}$ and $[\cdot;\cdot ]$ is the convolutional operation and concatenation operation, respectively. The $F_{avg}^s$ and $F_{max}^s$ of two-dimensional maps are generated through avgpooling and maxpooling. The spatial attention weight map $M_{s}(F')$ is used to select the informative pixels and as the input of the following downsampling layer.

The objective of the decoder is to accurately classify each pixel. Therefore, we utilize the cross entropy of the multi-class classification for each pixel as the loss function. For a batch of frames, the loss function of the entire LSSC system can be given by
\begin{equation}
    L_{LSSC} = -\frac{\sum\limits_{b=1}^{B}\sum\limits_{l=1}^{H_{l} \times W_{l}}  \sum\limits_{i=1}^{N}(p_{b,l,i}log(\widehat p_{b,l,i})}{S \times H_{l} \times W_{l}},
\end{equation}
where $B$ is the batch size, $H_{l}$ and $W_{l}$ is the input image feature of one layer. $\widehat p_{b,l, i}$ is the predicted probability that the pixel is classified as a category $i$ in the $b$ batch at layer $l$. $p_{b,l, i}$ is the classification indicator with a value of 0 or 1 and $N$ is the number of categories. By the loss function, the LSSC system is trained by the end-to-end methods.
\vspace{-0.3cm}

\begin{center}
\begin{table}[t]
\caption{Simulation Parameters}
\label{tab:my-table}
    \begin{tabular}{|c|c|c|c|}
    \hline
    \textbf{Module} &
      \textbf{Layer Name} &
      \textbf{Block} &
      \textbf{Value} \\ \hline
    \multirow{3}{*}{\begin{tabular}[m]{@{}c@{}}
    Semantic 
    \\
    Encoder\end{tabular}} &
      \multirow{3}{*}{3xDownsampling} &
      (ConvBN+ReLU)x2 &
      \begin{tabular}[m]{@{}c@{}}kernel size=3\\ stride=1\end{tabular} \\ \cline{3-4} 
     &
       &
      Maxpool(k=2,s=2) &
      \begin{tabular}[c]{@{}c@{}}kernel size=2\\ stride=2\end{tabular} \\ \cline{2-4} 
     &
      \multirow{3}{*}{CBAM} &
      \begin{tabular}[m]{@{}c@{}}AdaptiveAvgPool2d\\ AdaptiveMaxPool2d\end{tabular} &
      output\_size=1 \\ \cline{3-4} 
     &
       &
      \begin{tabular}[m]{@{}c@{}}Conv2d(k=7,s=1)\\ sigmoid\end{tabular} &
      \begin{tabular}[m]{@{}c@{}}kernel size=7\\ stride=1\end{tabular} \\ \hline
    AWGN &
      AWGN &
      SNR (dB) &
      1-20 \\ \hline
    \multirow{3}{*}{\begin{tabular}[m]{@{}c@{}}Semantic \\ Decoder\end{tabular}} &
      \multirow{3}{*}{3xUpsampling} &
      Upsampling &
      scale\_factor=2 \\ \cline{3-4} 
     &
       &
      (ConvBN+ReLU)x2 &
      \begin{tabular}[m]{@{}c@{}}kernel size=2\\ stride=1\end{tabular} \\ \hline
\end{tabular}
\end{table}
\end{center}
\vspace{-0.5cm}

\section{Simulation And Performance Analysis}
 
 To evaluate the performance of the proposed LSSC system, we use the dataset of Deep Automatic Portrait Matting for training and testing\cite{shen2016deep}. In our simulation, the coefficients $\lambda$, $\mu$, and $\nu$ that lead to the appropriate results are selected respectively 1, 0.1, and 0.2. We train the model using the training set from the image dataset, which contains 1530 training images and 170 test images, and every image is in RGB format. The learning rate $\eta$ is $10^{-3}$. The proposed LSSC system is trained under AWGN with the SNR varying randomly between 1 and 20 dB and the RMSprop algorithm is employed to optimize the model. The simulation parameters are listed in Table I. For comparison purposes, we use four baselines: a) UNet, b) Attention-UNet with an attention gate (AG) module proposed in \cite{oktay2018attention}, c) CBAM-UNet with a CBAM module in each downsampling process and d): the traditional modular wireless transmission mode, which adopts PNG for source coding, Low-Density Parity-Check Codes (LDPC) for channel coding, and QAM for modulation\cite{10118717}. At the receiver, we employ the UNet to get the received image segmentations. 

In Fig. 5, we show the relationship between the number of downsampling layers and the optimization objective $Q$ of our proposed model. From Fig. 5, we can see that when the number of downsampling layers is 3, the optimization objective $Q$ of the proposed model reaches the peak. When the number of downsampling layers is 2, the mIoU is lower which can not achieve high accuracy of image segmentation. And when the number of downsampling layers is 4, we can observe that the FLOPs and model size are bigger, which increases the burden on IoT devices. Fig. 5 demonstrates that in our proposed model architecture, adopting three number of downsampling layers can exploit the trade-off between accuracy and complexity, and can achieve the highest accuracy with limited calculations.

 In Tabel II, we show the FLOPs, model size, and mIoU of different schemes. From Table II, we can see that the FLOPs of our proposed model is about 13 $\%$ lower than UNet. Reducing the computational complexity of the semantic segmentation model makes it possible for IoT devices to run the semantic codec. Meanwhile, the model size of our proposed codec is about 55 $\%$ less than UNet, because of the decrease in the number of downsampling layers. Table II also indicates that, among the compared methods, the proposed codec in this paper, achieves the minimal computational complexity and lowest model size while maintaining high segmentation performance. Meanwhile, Table II also shows that the size of the CBAM module is almost negligible, which can improve the model segmentation performance.

 Fig. 6 shows how the mIoU changes with different channel SNR. We can see that as the channel SNR increases, the mIoU of both the LSSC system and the traditional schemes increase, and the LSSC system has a better performance at low SNR region because the semantic features contain contextual information. We can also observe that the mIoU of the LSSC system is more robust with the steadier mIoU, in contrast to the traditional schemes. The line of the traditional scheme sharply drops when the channel quality deteriorates because the damaged PNG image cannot be segmented. Besides, the reason for the steep line of the traditional scheme is that it encodes the image into flag and data bits, which would be transmitted incorrectly in the low SNR region. The wrong flag bits will severely compromise the image reconstruction and damage the accuracy of image segmentation, which is known as the ``cliff effect''\cite{bourtsoulatze2019deep}. From Fig. 6, we see that the ``cliff effect'' is avoided by the LSSC system. This is because the LSSC system can effectively extract image semantic features which contain contextual information, thus achieving more robust transmission against channel variation.

\begin{figure}[t]
\centering
\setlength{\belowcaptionskip}{-0.45cm}
\includegraphics[width=8cm]{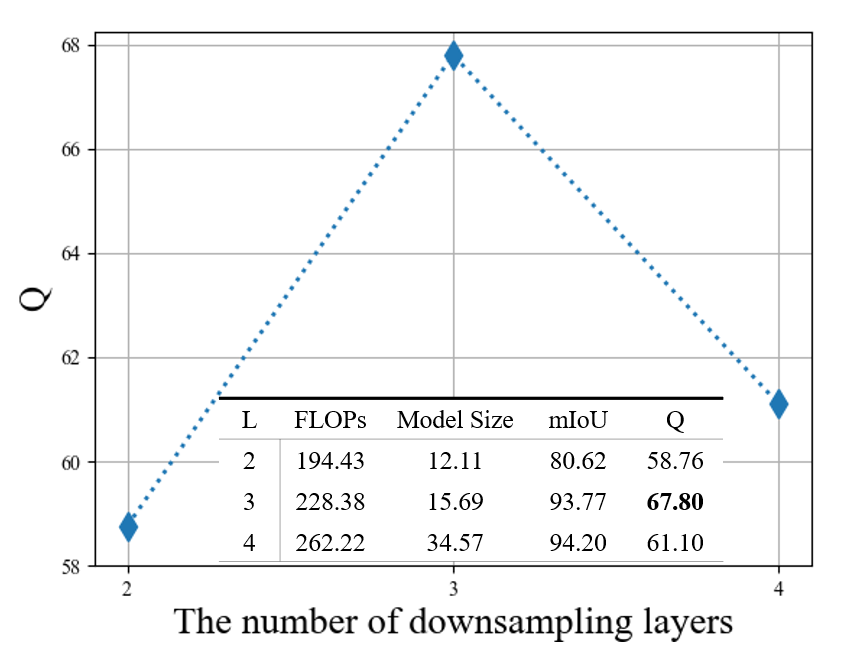}
\centering
\caption{The relationship between the number of downsampling layers and objective $Q$.}
\label{fig2}
\end{figure}

\begin{table}[t]
\caption{\textbf{The relationship between mIoU and FLOPs under LSSC system and different models}}
\centering
\begin{tabular}{cccc}
\toprule
Methods&FLOPs &Model Size &mIoU(\%) \\
\midrule
UNet           & 262.12          & 34.53          & 94.07    \\ 
Attention-UNet & 266.54          & 34.88          & 94.18    \\ 
CBAM-UNet      & 262.22          & 34.57          & 94.20    \\ 
\textbf{LSSC(Ours)}  & \textbf{228.38} & \textbf{15.69} & \textbf{93.77}  \\ 
\bottomrule
\end{tabular}
\end{table}
\vspace{-1em}
\begin{figure}[t]
\centering
\setlength{\belowcaptionskip}{-0.45cm}
\includegraphics[width=9cm,height=7cm]{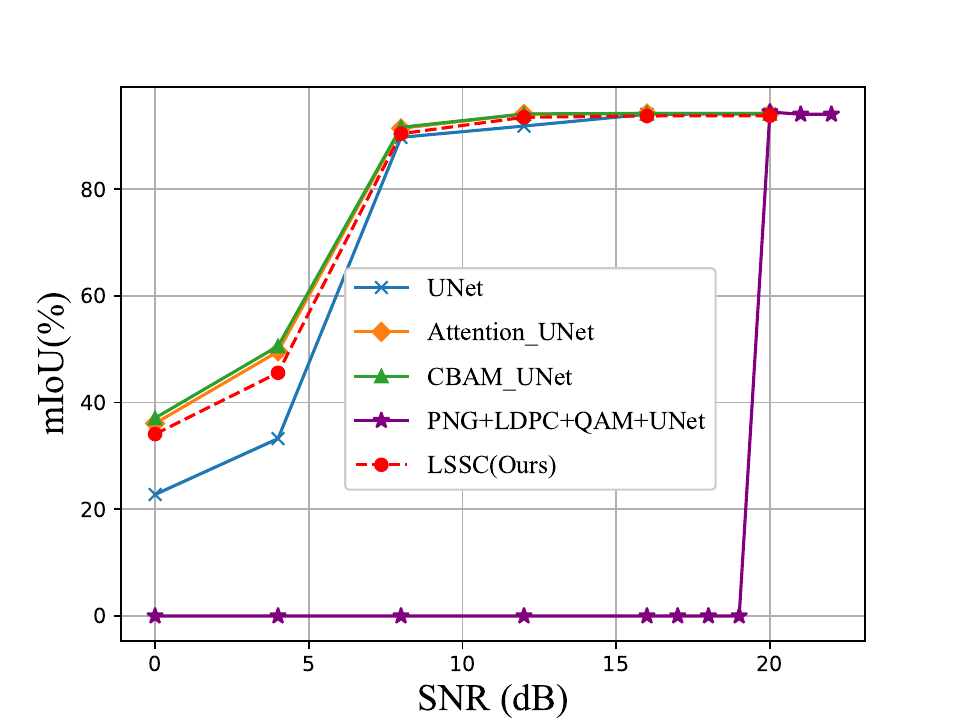}
\centering
\caption{mIoU comparison of the LSSC system.}
\label{fig2}
\end{figure}
	
\section{Conclusion}
In this paper, we proposed an attention-based UNet enabled lightweight image communication (LSSC) system over the Internet of Things, which has lower computational complexity and smaller model size. To lessen the burden of IoT devices, the LSSC system transmits the trained semantic codec and semantic features between the edge server and IoT devices. Specifically, by employing the CBAM module in the semantic encoder, the proposed LSSC codec efficiently learns informative image information by identifying the most important channels and pixels. Simulation results show that compared to the traditional UNet, our system achieves lower FLOPs and smaller model size with a sacrifice of 3\% accuracy. Moreover, the LSSC system performs better in the low SNR region than the traditional communication scheme. Therefore, our proposed LSSC system is a promising candidate for IoT devices, especially in the low SNR region. 

\bibliographystyle{IEEEbib}
\renewcommand{\baselinestretch}{0.7}
\bibliography{BNN}
\end{document}